\documentclass[10pt,twocolumn,letterpaper]{article}

\usepackage{amssymb}
\usepackage{pifont}
\usepackage[pagenumbers]{cvpr} 

\definecolor{cvprblue}{rgb}{0.21,0.49,0.74}
\usepackage[pagebackref,breaklinks,colorlinks,allcolors=cvprblue]{hyperref}

\def\paperID{xxxx} 
\def\confName{CVPR}
\def\confYear{2026}

\newcommand{\name}{EgoReAct\xspace}
\newcommand{\dbname}{HRD\xspace}

\title{\name: \underline{Ego}centric Video-Driven 3D Human \underline{React}ion Generation}

\author{
Libo Zhang\textsuperscript{1,$\S$} \quad
Zekun Li\textsuperscript{2} \quad
Tianyu Li\textsuperscript{3} \quad
Zeyu Cao\textsuperscript{4} \quad
Rui Xu\textsuperscript{5} \quad
Xiaoxiao Long\textsuperscript{6} \quad
\\
Wenjia Wang\textsuperscript{5} \quad
Jingbo Wang\textsuperscript{7} \quad
Yuan Liu\textsuperscript{8} \quad
Wenping Wang\textsuperscript{9} \quad
Daquan Zhou\textsuperscript{10} \quad
\\
Taku Komura\textsuperscript{5} \quad
Zhiyang Dou\textsuperscript{5,11,$\dagger$}
\\[0.6em]
\textsuperscript{1}THU \quad
\textsuperscript{2}Brown \quad
\textsuperscript{3}Georgia Tech \quad
\textsuperscript{4}Cambridge \quad
\textsuperscript{5}HKU \quad
\textsuperscript{6}NJU \quad
\\
\textsuperscript{7}CUHK \quad
\textsuperscript{8}HKUST \quad
\textsuperscript{9}TAMU \quad
\textsuperscript{10}PKU \quad
\textsuperscript{11}MIT
}

\begin{document}
\newcommand{\teaserCaption}{\name takes streaming egocentric video as input and synthesizes spatially grounded, realistic human reaction motions in real time, enabling responsive full-body behaviors that are tightly coupled with the ongoing egocentric observations.}

\twocolumn[{
    \renewcommand\twocolumn[1][]{#1}
    \maketitle
    \centering
    \vspace{-2mm}
    \begin{minipage}{1\textwidth}
        \centering 
        \includegraphics[width=\textwidth]{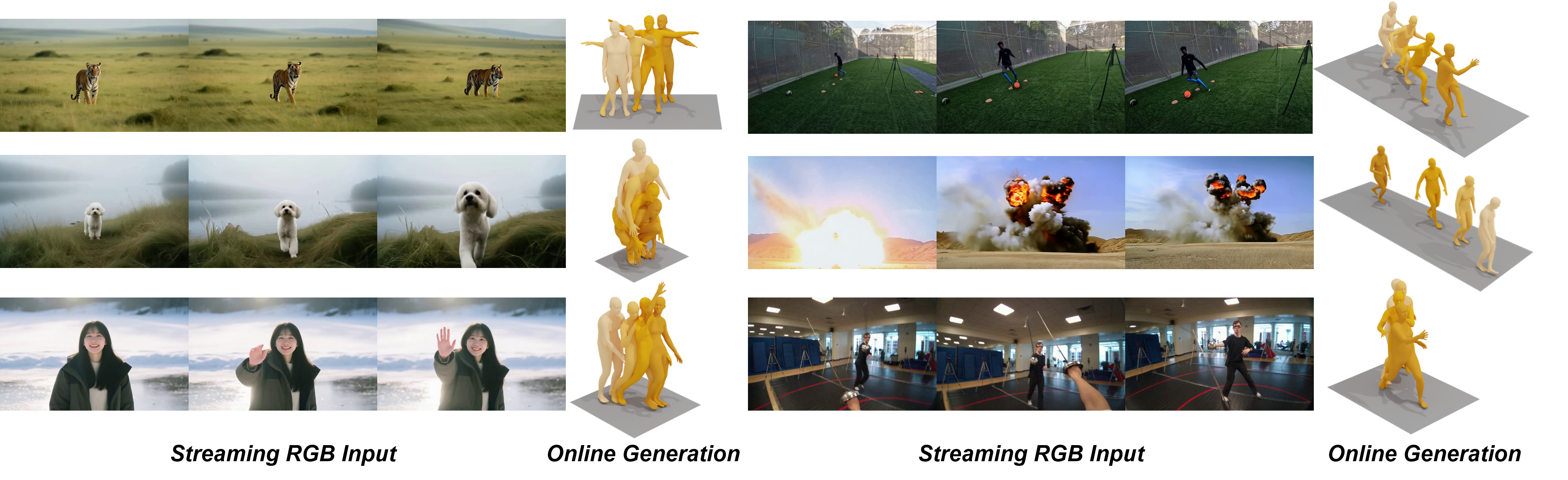}
    \end{minipage}
    
    \captionsetup{type=figure}
    \vspace{-2mm}
    \captionof{figure}{\teaserCaption 
    }
    \label{fig:teaser}
    \vspace{3mm}
}]

\maketitle

\begingroup
\renewcommand\thefootnote{}
\footnotetext{
\noindent
$^{\S}$, $^{\dagger}$ denote work completed during Summer 2025 and corresponding authors, respectively.
}
\endgroup

\begin{abstract}
Humans exhibit adaptive, context-sensitive responses to egocentric visual input. 
However, faithfully modeling such reactions from egocentric video remains challenging due to the dual requirements of strictly causal generation and precise 3D spatial alignment. 
To tackle this problem, we first construct the Human Reaction Dataset (\textbf{\dbname}) to address data scarcity and misalignment by building a spatially aligned egocentric video–reaction dataset,
as existing datasets (e.g., ViMo) suffer from significant spatial inconsistency between the egocentric video and reaction motion, e.g., dynamically moving motions are always paired with fixed-camera videos. Leveraging \dbname, we present \textbf{\name}, the first autoregressive framework that generates 3D-aligned human reaction motions from egocentric video streams in real-time. We first compress the reaction motion into a compact yet expressive latent space via a Vector Quantised-Variational AutoEncoder and then train a Generative Pre-trained Transformer for reaction generation from the visual input. \name incorporates 3D dynamic features, i.e., metric depth, and head dynamics during the generation, which effectively enhance spatial grounding. Extensive experiments demonstrate that \name achieves remarkably higher realism, spatial consistency, and generation efficiency compared with prior methods, while maintaining strict causality during generation.
We will release code, models, and data upon acceptance.
\vspace{-4mm}
\end{abstract}
    
\section{Introduction}
\label{sec:intro}

Faithful modeling human behavior in response to various input conditions like environmental dynamics remains a long-standing and fundamental challenge, with broad implications for robotics~\cite{lee2023locomotion, react_to_this, Heuer2024Benchmarking}, computer vision~\cite{bhatnagar2022behave, sem_geo_mo, gaze_prediction}, and computer graphics~\cite{jiang2024autonomous, Yao2025HOSIG}. 
Different from previous efforts in human motion generation that conditioned on modalities like textual instructions~\cite{Tevet2023HumanMotionDiffusion, t2m-gpt, guo2023momask,zhou2024emdm}, another person’s motion~\cite{Chopin2022InteractionTransformer, Liang2024InterGen, Tan2025ThinkThenReact, wu2025text2interact}, or spatial audio~\cite{xu2025mospa}.
This paper focuses on high-quality reaction motion synthesis with ego-centric perception, \ie RGB streaming video\footnote{By RGB streaming video, we emphasize \textit{causal, per-frame} processing, i.e., the model has no access to future context.}, as humans inherently perceive and react to the world from their own first-person viewpoint to a large extent--one of the most intuitive and general paradigm for modeling reactions.

However, collecting large-scale real world ego-centric based reaction data is inherently difficult, 
as it requires precise synchronization between camera motion and the actor’s body movement, 
especially the head trajectory which is a principal cue for the egocentric view. 
This remains a major obstacle in learn a model to generate
human reactions from ego-centric perception. A very recent effort, the ViMo dataset~\cite{yu2025hero}, provides manually paired videos and human reaction motions. 
However, the quality and diversity of the data are highly limited, which restricts their effectiveness for model training.
Most videos are from static cameras even though the body is moving, which deviates from true egocentric perception where viewpoint changes are induced by self-motion; 
this mismatch leads to ambiguity and poor spatial alignment; 
See Fig.~\ref{fig:dataset_examples}. 
In this paper, we create a more spatially aligned dataset named \dbname through an automatic data generation pipeline (\cref{subsec: dataset}).
This scalable synthesis strategy enables the creation of a large, diverse, and spatially aligned dataset that supports data-driven learning of realistic, spatially grounded reactions. 


Leveraging \dbname, we introduce \name, the first autoregressive framework for this task, capable of generating spatially grounded human reaction motions from egocentric videos in real time. 
To achieve both efficient and high-quality 3D motion synthesis, we formulate the generation process within a causal autoregressive paradigm.
Specifically, we adopt a two-stage design.
In the first stage, a VQ-VAE is learned to discretize continuous motion sequences into compact code indices.
In the second stage, we train an autoregressive Transformer to predict the code index of the current motion frame based on historical context and the latest visual observation.
This design enables the model to generate reaction motions sequentially and causally, updating its predictions as new visual observations arrive in a streaming manner. Note that prior work~\cite{yu2025hero} assumes access to the entire video—including future frames—to one-shot generate the reaction sequence, which violates causality in egocentric settings requiring online synthesis.


Instead of relying solely on 2D visual semantics from egocentric videos, which neglects the crucial spatial alignment between the generated reaction motions and the ego perception, we further enhance the generation process with 3D dynamic features by leveraging the power of vision foundation models. Specifically, given a steaming RGB video, 
our method first extracts rich 2D visual semantics using a pretrained foundation vision model~\cite{oquab2023dinov2}, 
which guide the autoregressive generation of plausible human reactions.
Building upon this 2D perception, we further incorporate 3D dynamic cues—including the metric depth estimated from the video~\cite{video_depth_anything} and head dynamics as an ego-motion signal—to achieve spatially grounded and geometrically consistent motion synthesis.
These complementary geometric signals enable the synthesized motions to remain consistent with the egocentric visual context within a unified 3D coordinate frame.

Based on qualitative and quantitative results, our framework produces realistic, time-consistent, and spatially grounded human reactions that more faithfully emulate human perceptual behavior in dynamic real-world environments, achieving \textbf{0.456} vs. 0.661 on FID and \textbf{21.33ms} vs 400.56 ms on first frame latency compared to the SOTA baseline~\cite{yu2025hero}. Our contributions are summarized as follows:
\begin{itemize}
    \item We construct HRD dataset that contains diverse and spatially aligned egocentric video–reaction motion pairs, providing geometry-consistent supervision for training and evaluation.
    \item We propose EgoReAct, the first autoregressive framework for real-time, 3D-aligned human reaction generation from egocentric videos.
    \item Comprehensive experiments demonstrate the effectiveness of our framework, achieving more efficient and spatially grounded reaction motion generation. 
\end{itemize}

\section{Related Work}
\label{sec:related}
\begin{figure*}[t]
     \centering
     \vspace{-2mm}
    \includegraphics[width=0.97\textwidth]{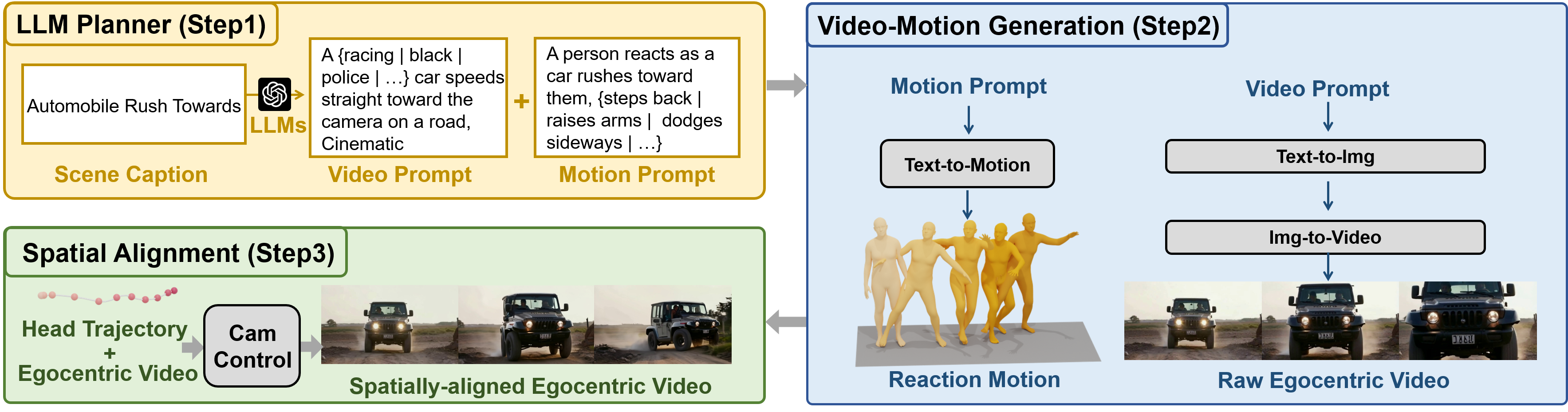}
        \vspace{-2mm}
    \caption{\textbf{The automated pipeline for generating the Spatially Aligned Human Reaction Dataset (HRD).} Given a scene caption, we first employ LLMs to produce video and motion prompts, then generate egocentric videos and reaction motions through text-driven generation, followed by spatial alignment via camera trajectory control.}
    \label{fig:dataset_gen_pipeline}
        \vspace{-4mm}
\end{figure*}

\subsection{Conditional Motion Generation}
Conditional motion generation aims to synthesize human motions guided by external signals such as text~\cite{Tevet2023HumanMotionDiffusion, t2m-gpt, zhou2024emdm, guo2023momask, wan2024tlcontrol, rmd, lu2025scamo, fan2025go, cong2024laserhuman, wan2024tlcontrol}, audio~\cite{controllable_group, tseng2023edge, listen_denoise_action, xu2025mospa}, action~\cite{guo2020action2motion, xu2023actformer, chen2024pay}, motion history~\cite{feng2024motionwavelet, chen2023humanmac, sun2023towards, wei2023human, barquero2023belfusion}, etc.
Among various conditioning modalities, text-to-motion has been most extensively explored.
Human Motion Diffusion Model~\cite{Tevet2023HumanMotionDiffusion} first introduce diffusion models into motion synthesis, achieving high realism and temporal coherence but at the cost of slow sampling. 
EMDM~\cite{zhou2024emdm} further improve efficiency by accelerating the diffusion process, enabling fast yet high-quality motion generation.
T2M-GPT~\cite{t2m-gpt} reformulate this task in a discrete latent space using VQ-VAE and a GPT-style transformer, enabling autoregressive text-conditioned motion generation with improved semantic alignment.
Furthermore, recent approaches~\cite{guo2023momask, javed2024intermask} leverage masked motion modeling to achieve fast and high-quality motion generation by exploiting efficient inference and bidirectional attention mechanisms.
Beyond language, researchers have investigated motion generation conditioned on other modalities, including audio cues~\cite{tseng2023edge, listen_denoise_action, xu2025mospa}, and action-conditioned human-human interaction~\cite{Chopin2022InteractionTransformer, Liang2024InterGen, Tan2025ThinkThenReact}.
These methods have shown promising results in modeling reaction motions conditioned on spatial audio cues or the dynamic actions of surrounding agents.
However, these modalities remains limited for comprehensive reaction modeling.
In contrast, ego-perception--the first-person visual sensing--provides a more complete and direct understanding of how humans perceive and react to their environment.
Motivated by this insight, our work focuses on Ego-Driven Reaction Generation, where reactions are synthesized directly from the dynamic egocentric visual stream.
\subsection{Ego-Driven Reaction Generation}
Egocentric visual inputs can be leveraged for a variety of purposes in human motion understanding.
A series of recent works~\cite{Li2023EgoBodyPose, wang2024egocentric, yi2025egoallo, Wang2025Ego4o, pan2025lookout} aim to estimate human body pose from egocentric videos.
Beyond pose estimation, egocentric observations have also been used to drive humans to perform tasks, bridging perception and action~\cite{moving_by_looking}.
However, the aforementioned works do not establish an connection between the dynamics of the surrounding world and the resulting human  behaviors.
Recently, a few works~\cite{xiu2025egotwin, patel2025uniegomotion} have attempted to model the relationship between human motion and egocentric perception using video diffusion models.
Nevertheless, modeling human reactions directly from egocentric perception remains underexplored. 
The most related work is HERO~\cite{yu2025hero}, which generates human reactions from RGB egocentric videos.
However, it overlooks both the spatial constraints of ego perception and the fact that a reaction model should be casual.
To this end, we propose a autoregressive reaction generation model driven by egocentric videos, which grounds both perception and motion generation within the same physical 3D space.


\begin{figure*}[t]
    \centering
    \includegraphics[width=\textwidth]{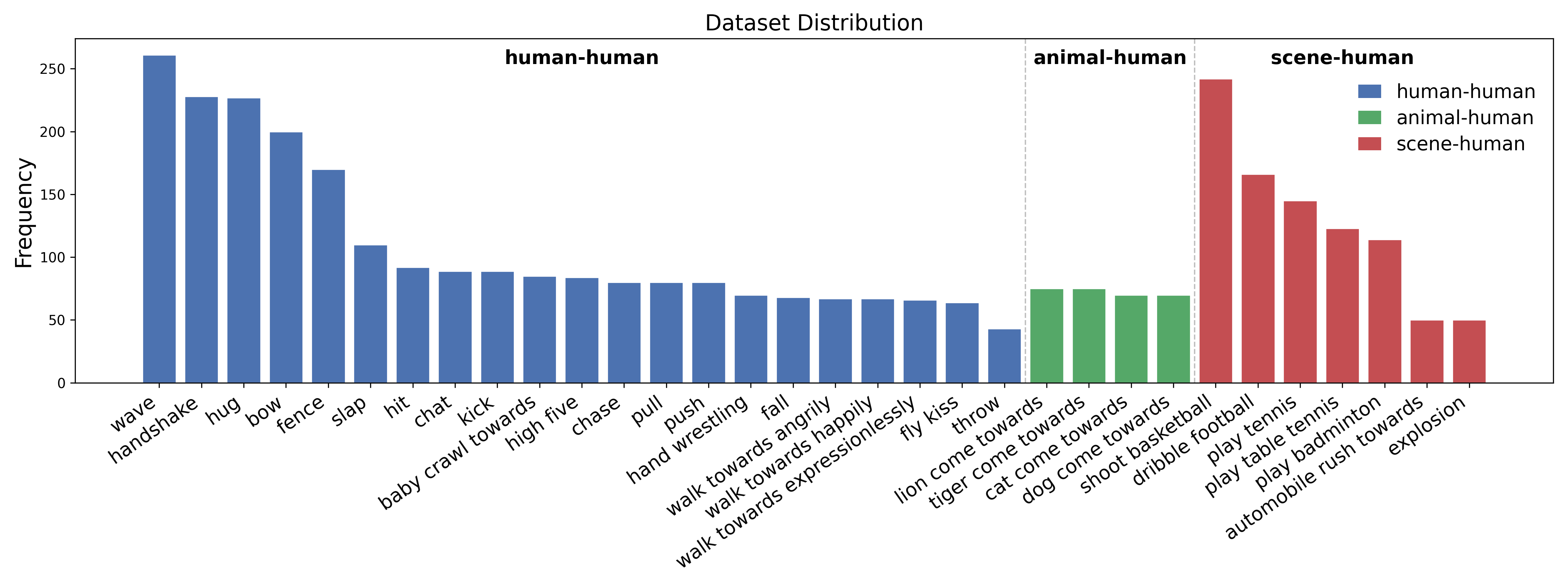}
    \vspace{-6mm}
    \caption{\textbf{Dataset Distribution of the \dbname dataset.} The dataset consists of three main categories: human-human (blue), animal-human (green), and scene-human (red) interactions.
    }
    \label{fig:data_dist}
    \vspace{-4mm}
\end{figure*}

\section{Dataset}
\label{subsec: dataset}
Existing datasets for egocentric reaction generation often suffer from spatial misalignment between the camera view and the corresponding human motion.
To address this limitation, we build a spatially aligned egocentric video–human reaction dataset (\dbname) via a scalable, fully automated generation pipeline, as illustrated in Fig.~\ref{fig:dataset_gen_pipeline}.
Given a category caption describing the scene observed from a first-person perspective, our pipeline automatically generates the corresponding egocentric video and its paired reaction motion, which are spatially-aligned.
Specifically, we first employ a Large Language Model~\cite{openai_chatgpt} to generate textual descriptions for both the egocentric video and the corresponding reaction motion based on the given caption.
For the egocentric video, we use the textual descriptions to synthesize realistic and temporally coherent first-person videos with a text-to-image model~\cite{stable_diffusion, sdxl, qwen} followed by an image-to-video diffusion model~\cite{kuaishou_kling_2024, alibaba_wan_2025, google_veo_2025}.
For the reaction motion, we leverage a state-of-the-art text-to-motion model~\cite{guo2023momask} to generate the corresponding human reactions.
We then extract the head trajectory from the reaction motion as the camera trajectory, 
and use recent camera-controlled video generation approaches~\cite{trajectorycrafter, ex4d, recammaster} to synthesize new egocentric videos based on the updated trajectory and the original video, 
ensuring spatial alignment between the video and human motion.

\begin{figure*}[t]
    \centering
        \vspace{-3mm}
    \includegraphics[width=\textwidth]{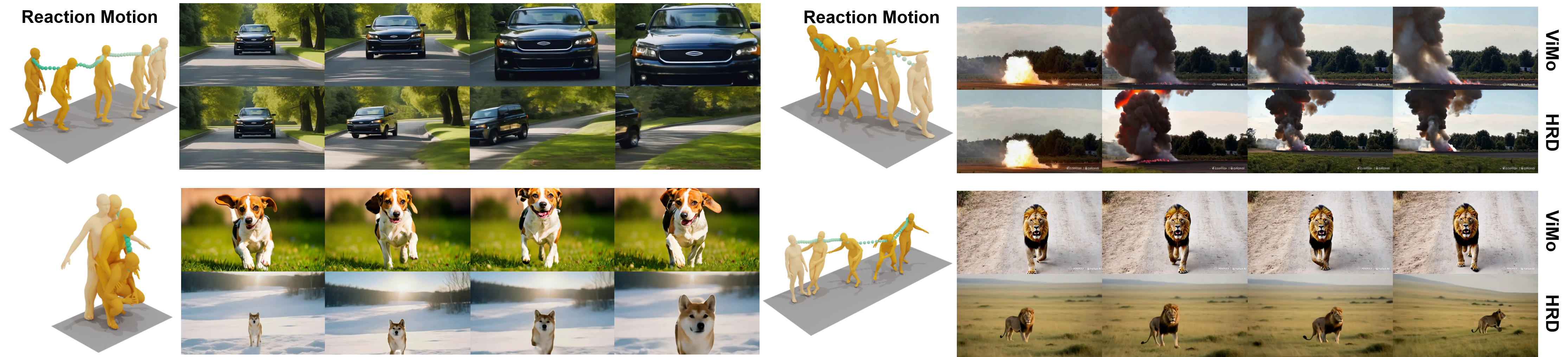}
    \caption{\textbf{Comparison between ViMo~\cite{yu2025hero} and our Spatially Aligned Human Reaction Dataset (HRD).} 
    The left side shows the ground-truth reaction motion. On the right, the top row presents the egocentric video from ViMo, while the bottom row shows the video from our HRD. 
    Our dataset provides significantly more accurate spatial alignment between the egocentric video and the reaction motion.}
    \label{fig:dataset_examples}
    \vspace{-2mm}
\end{figure*}

To better evaluate the limitations of existing datasets containing egocentric videos and reaction motions, 
we use our automated pipeline to fix the previously proposed ViMo dataset~\cite{yu2025hero} by re-synthesizing spatially aligned egocentric videos based on the corresponding reaction motions.
The resulting \dbname dataset contains 3,500 spatially aligned egocentric video–motion pairs across 32 diverse perceptual categories, including automobile rush towards, shooting basketball, and other scenarios, with each clip lasting about 150 frames. 
All data undergoes expert manual review to ensure quality and consistency.
Several examples are shown in Fig.~\ref{fig:dataset_examples}, where the original ViMo samples exhibit spatial misalignment between the egocentric videos and motions, while our automated data generation pipeline produces much more spatially consistent results, which provides 3D-aware supervision for both training and evaluation of reaction generation models. 
The distribution of the \dbname dataset is further illustrated in Fig.~\ref{fig:data_dist}, where we present the different motion categories and their corresponding frequencies.

\section{Method}
Given an input RGB video $V=\{I_1, I_2, ..., I_T\}$ where $I\in\mathbb{R}^{H\times W\times 3}$ and $T$ is the number of frames, our goal is to generate the temporally and spatially aligned 3D human reaction motion $M\in \mathbb{R}^{T\times D}$ in a streaming manner, where $D$ is the dimension of human pose.
Our framework consists of two main modules: Ego Perception Representation Modeling (Sec~\ref{subsec: Encoding}) and Auto-regressive Human Reaction Model (Sec~\ref{subsec: Auto-regressive}), which is illustrated in Fig.~\ref{fig:overview}.
The former encodes both visual semantics and dynamic information from the egocentric video, along with geometric features, into a latent space. 
The latter then auto-regressively generates the corresponding human reaction motion conditioned on the encoded features, ensuring spatial consistency and temporal causality across the sequence.
\subsection{Ego Perception Representation Modeling}
\label{subsec: Encoding}
Previous works~\cite{yu2025hero} mainly rely on 2D visual cues from egocentric videos, making it difficult to generate 3D-aware reaction motion.
In contrast, we incorporate both visual and dynamic information from videos, along with geometric inputs, to produce spatially consistent motion generation.
Moreover, rather than conditioning the model on the entire video, which disrupts causality in ego-perception–driven human reaction generation, we reformulate the conditional encoding mechanism introduced in HERO~\cite{yu2025hero}.

\noindent\textbf{Ego Perception Representation.}
Given a streaming video sequence ${I_t}_{\{1:N\}}$, our goal is to model a robust representation that is semantically rich, spatially grounded, and temporally causal.
To this end, we extract three complementary modalities from each frame $I_t$: visual semantics $f_s$, ego-stereo depth $f_d$, and head dynamics $f_h$, and encode them into a shared latent space.
For each input frame $I_t \in \mathbb{R}^{H\times W\times 3}$, we employ DINOv2~\cite{oquab2023dinov2} as the visual encoder and use the final-layer \texttt{[CLS]} token as the semantic feature $f_s \in \mathbb{R}^{384}$.
To enhance spatial awareness, we estimate the metric depth $D_t \in \mathbb{R}^{H\times W}$ of each frame using a recent video depth estimation method~\cite{video_depth_anything}, and further encode it through a shallow convolutional network to obtain the ego-stereo feature $f_d \in \mathbb{R}^{384}$.
Moreover, to maintain spatial alignment and causal consistency in ego-perception–driven reaction modeling, we incorporate the previous-frame head velocity $V_{t-1}$ in 3D space as an additional conditioning signal $f_h \in \mathbb{R}^{384}$, which is embedded into the latent space via a two-layer MLP with 512 hidden units.
During inference, the head velocity $V_{t}$ is computed by decoding the full motion from the previously predicted motion token and calculating the difference in head position between consecutive frames:
\begin{equation}
    V_t = \frac{P_{t} - P_{t-1}}{\Delta t},
\end{equation}
where $P_t\in\mathbb{R}^3$ and $P_{t-1}\in\mathbb{R}^3$ represent the 3D head positions at frames $t$ and $t-1$, respectively, and $\Delta t$ is the time interval between the two frames.
For the first and second frames, we initialize the head velocity to 0, as there is no previous motion to reference.

By encoding visual-geometry-dynamic information from the input streaming video and head dynamics into a shared latent space, our ego-perception representation enable our model to achieve a more 3D-aware and temporally consistent reaction motion generation.

\begin{figure*}[t]
    \vspace{-3mm}
    \centering
    \includegraphics[width=0.97\textwidth]{imgs/method_pipeline.png}
        \vspace{-2mm}
    \caption{\textbf{Pipeline of EgoReAct.} We first learn a Motion VQ-VAE to discretize continuous 3D motions into compact token sequences. Building on this representation, EgoReAct takes streaming egocentric RGB frames as input, estimates their depth, and encodes the image, depth, and head dynamics cues to form ego-perception features, which guide an Autoregressive Transformer to sequentially generate spatially aligned and temporally causal reaction motions.
    }
    \label{fig:overview}
        \vspace{-3mm}
\end{figure*}

\noindent\textbf{Token Fusion for Autoregressive Generation.}
In our framework, we fuse multiple modalities, such as visual semantics, depth, head dynamics, 
and the previously predicted motion ID (with the first motion ID initialized as a learnable token \texttt{[BOS]}) into a unified fusion token for each frame,
which serves as the conditioning signal for the autoregressive generation process.
Specifically, we concatenate these features $f_s\in\mathbb{R}^{384}$, $f_d\in\mathbb{R}^{384}$, $f_h\in\mathbb{R}^{384}$, $f_m\in\mathbb{R}^{384}$ (where $f_m$ represents the motion ID embedding) to form a concatenated vector:
\begin{equation}
    f_{\text{fusion}} = \text{CONCAT} (f_s, f_d, f_h, f_m) \in \mathbb{R}^{1536},
\end{equation}
which is then passed through a linear layer with 512 hidden units to produce a unified token representation:
\begin{equation}
    f_{\text{token}} = \text{MLP}(f_{\text{fusion}}) \in \mathbb{R}^{384}.
\end{equation}
This fusion token is then fed into the autoregressive Transformer, which generates the human reaction motions in a sequential manner, ensuring both spatial alignment and temporal coherence.
By combining rich visual, dynamic, and geometric cues into a single token, we allow the model to focus on the most informative aspects of the input for motion synthesis, leading to more spatially consistent and temporally smooth outputs.
\subsection{Auto-regressive Human Reaction Model}
\label{subsec: Auto-regressive}
Previous approaches~\cite{yu2025hero} generate the entire reaction motion after observing the full egocentric video sequence, which is impractical for real-world applications.
We instead design an auto-regressive human reaction model that predicts motion frame by frame based on past motion and the current perceptual input.
Our model consists of two components: a motion VQ-VAE that discretizes continuous motion sequences into a compact token space, and an auto-regressive Transformer that predicts motion tokens sequentially.

\noindent\textbf{Motion VQ-VAE.}
The motion VQ-VAE~\cite{humanml3d} encodes each motion sequence into a discrete latent space.
Given a ground-truth motion $M\in\mathbb{R}^{T\times D}$ (where $D=263$ following the HumanML3D~\cite{humanml3d} representation), the encoder $\mathbf{E}$ maps it to a continuous latent feature space $\mathbf{z}\in\mathbb{R}^{T\times d_c}$, where $d_c$ denotes the dimension of the latent feature.
The latent features are then quantized via a learnable codebook $\mathbf{C}=\{\mathbf{c}_k\}_{k=1}^K$ to obtain discrete embeddings $\mathbf{z}_q = \mathrm{Q}(\mathbf{z})\in\mathbb{R}^{T\times d_c}$,
where $\mathrm{Q}(\cdot)$ denotes the vector quantization operation.
The decoder $\mathbf{D}$ reconstructs the motion $\hat{M}=D(\mathbf{z}_q)$, enabling a compact and expressive discrete motion token space for the subsequent generative model. 
The VQ-VAE is optimized using the following loss function:
\begin{equation}
    \mathcal{L}_{vq} = ||M-\hat{M}||_1 + ||sg[\mathbf{z}]-\mathbf{z}_q||_2^2 + \beta||\mathbf{z}-sg[\mathbf{z}_q]||_2^2,
\end{equation}
where $sg[\cdot]$ denotes the stop-gradient operation, and $\beta$ is a hyperparameter that controls the weight of the commitment loss.

\noindent\textbf{Autoregressive Transformer.} 
With a learned motion VQ-VAE, a motion sequence $M=[m_1, m_2, ..., m_T]$ can be mapped to a sequence of discrete indices $S=[s_1, s_2, ..., s_{T/l}]$ through codebook quantization, where $l=4$ denotes the temporal downsampling rate.
The reaction generation task is then formulated as an autoregressive next-index prediction: 
$p(S_t|S_{<t}, H_{<t}, I_{\leq t}, D_{\leq t})$, where $H_{<t}, I_{\leq t}, D_{\leq t}$ represent the observed head dynamics, RGB images, and metric depth maps available up to time $t$, respectively.

To model this conditional probability, we adopt an auto-regressive Transformer~\cite{vaswani2017attention} that predicts each motion token $S_t$ conditioned on previous tokens and perceptual inputs,
where a causal mask is applied in the self-attention computation to prevent access to future information.
Denoting the likelihood of the full sequence as
$
P(S|H, I, D) = \prod_{t=1}^TP(S_t|S_{<t}, H_{<t}, I_{\leq t}, D_{\leq t})
$,
We maximize the log-likelihood of the data distribution:
\begin{equation}
    \mathcal{L}_{trans}=E_{S\sim P(S)}[-\text{log}P(S|H,I,D)].
\end{equation}

Overall, the autoregressive Transformer achieves frame-level control and real-time reaction generation by leveraging past motion tokens and current perceptual cues,
maintaining temporal coherence and 3D spatial consistency throughout the generated motion sequence.

\section{Experiments}
\label{sec:exp}

\begin{table*}[t]
\centering
\small
\setlength{\tabcolsep}{6pt}
\resizebox{0.9\linewidth}{!}{
\begin{tabular}{lcccccc}
\toprule
\textbf{Method} & \textbf{FID}$\downarrow$ & \textbf{Diversity}$\rightarrow$ & \textbf{MModality}$\uparrow$ & \textbf{Head Traj Error (cm)}$\downarrow$ & \textbf{FFL (ms)}$\downarrow$ & \textbf{Causal}\\
\midrule
Real & - & $8.098^{\pm0.054}$ & - & - & - & -\\
\midrule
MDM~\cite{tevet2022human} & $1.782^{\pm0.032}$ & $7.419^{\pm0.068}$ & $2.015^{\pm0.062}$ & $88.3^{\pm0.553}$ & 12230.53 & \ding{55} \\
EMDM~\cite{zhou2024emdm} & $1.275^{\pm0.021}$ & $7.572^{\pm0.079}$ & $1.785^{\pm0.059}$ & $81.6^{\pm0.540}$ & 81.27 & \ding{55} \\
HERO~\cite{yu2025hero} & $0.560^{\pm0.017}$ & $7.625^{\pm0.066}$ & $1.249^{\pm0.039}$ & $72.3^{\pm0.204}$ & 400.56 & \ding{55} \\
\textbf{Ours} & $\textbf{0.456}^{\pm0.016}$ & $\textbf{8.042}^{\pm0.053}$ & $\textbf{2.581}^{\pm0.060}$ & $\textbf{65.7}^{\pm0.397}$ & \textbf{21.33} & \ding{51}\\
\bottomrule
\end{tabular}
}
    \vspace{0mm}
\caption{\textbf{Quantitative evaluation on \dbname test set.}$\pm$ denotes the 95\% confidence interval, and $\rightarrow$ indicates the closer to the real motions the better.
Bold numbers highlight the best performance.}
\label{tab:comparison}
    \vspace{-3mm}
\end{table*}

\begin{figure*}[t]
    \vspace{-3mm}
    \centering
    \includegraphics[width=1.0\textwidth]{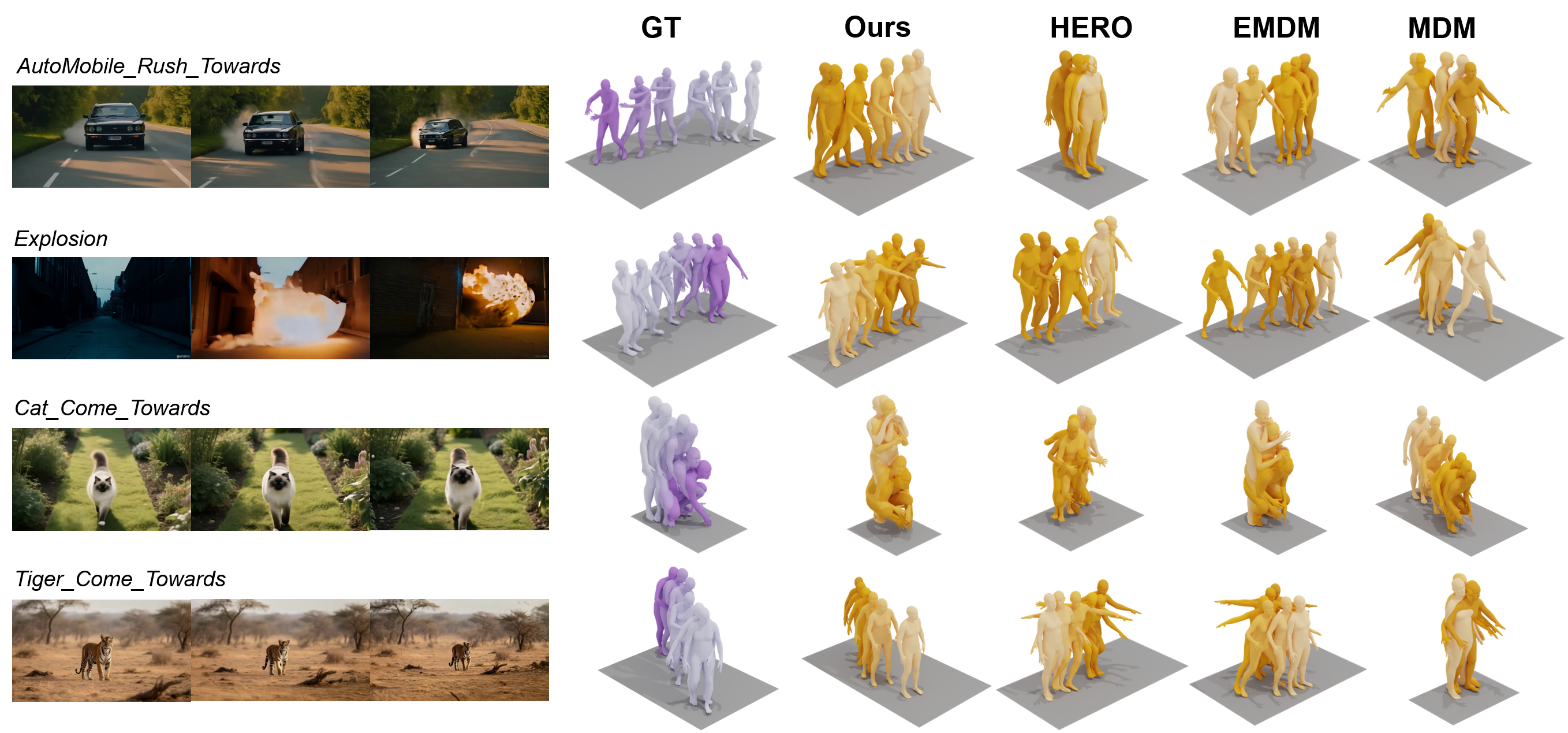}
        \vspace{-6mm}
\caption{\textbf{Qualitative Comparisons.} Our approach delivers stronger 3D grounding and motion plausibility—exhibiting improved spatial awareness, decisive reactions to egocentric observation, and fine-grained adjustments to near-field triggers—consistently outperforming HERO, EMDM, and MDM in the scenarios shown. The prediction results and GTs are in yellow and purple, respectively.}
    \label{fig:Qualitative}
    \vspace{-3mm}
\end{figure*}

\begin{figure}[t]
    \centering
    \includegraphics[width=0.96\linewidth]{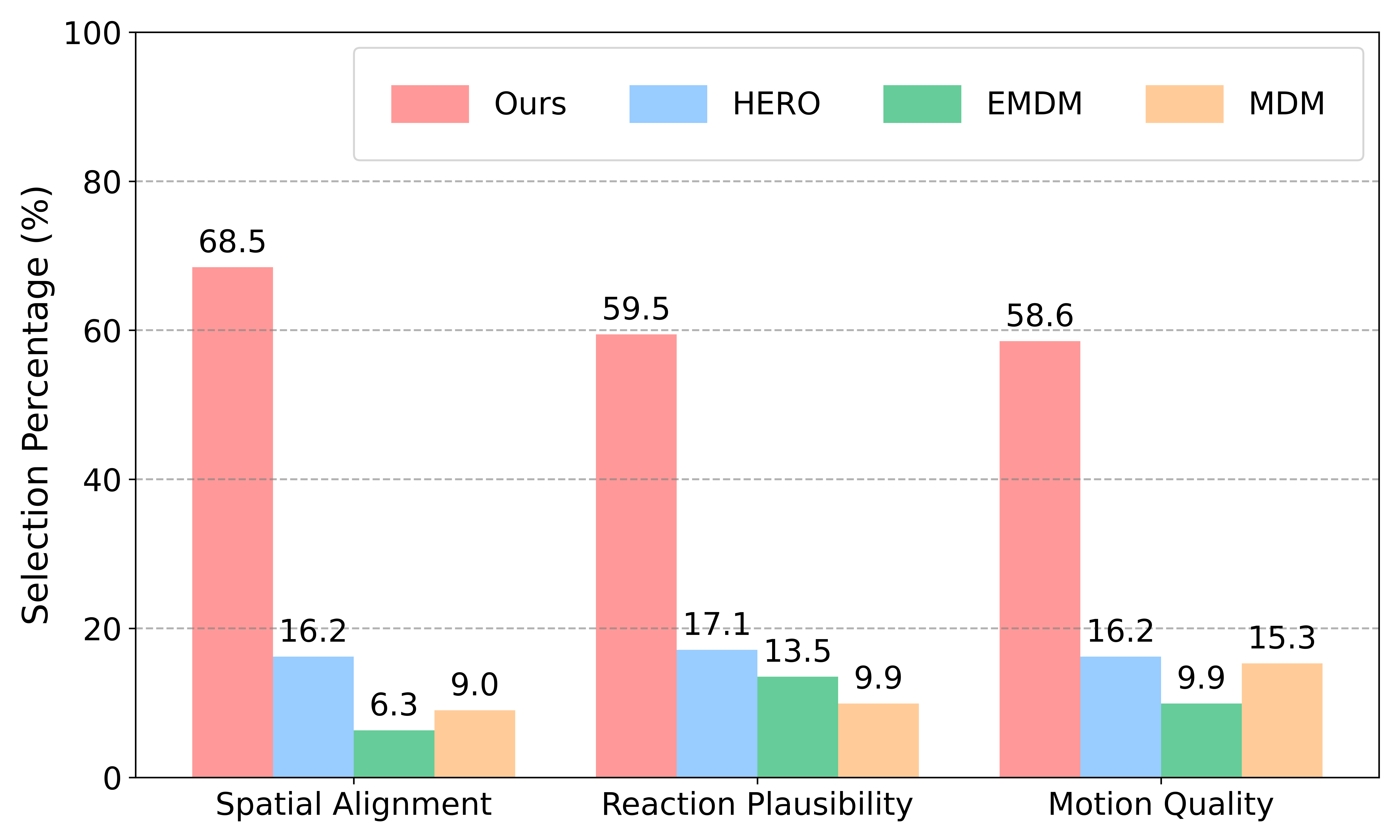}
    \vspace{-3mm}
\caption{\textbf{User study results.} Percentage of participant selections (higher is better) for three criteria—\emph{Spatial Alignment}, \emph{Reaction Plausibility}, and \emph{Motion Quality}—across competing methods. Our approach receives a clear majority of votes on all axes, indicating superior spatial grounding, more plausible reactions, and overall higher motion quality.}    
    \label{fig:user_study}
    \vspace{-4mm}
\end{figure}

\begin{figure}[t]
    \centering
    \includegraphics[width=0.96\linewidth]{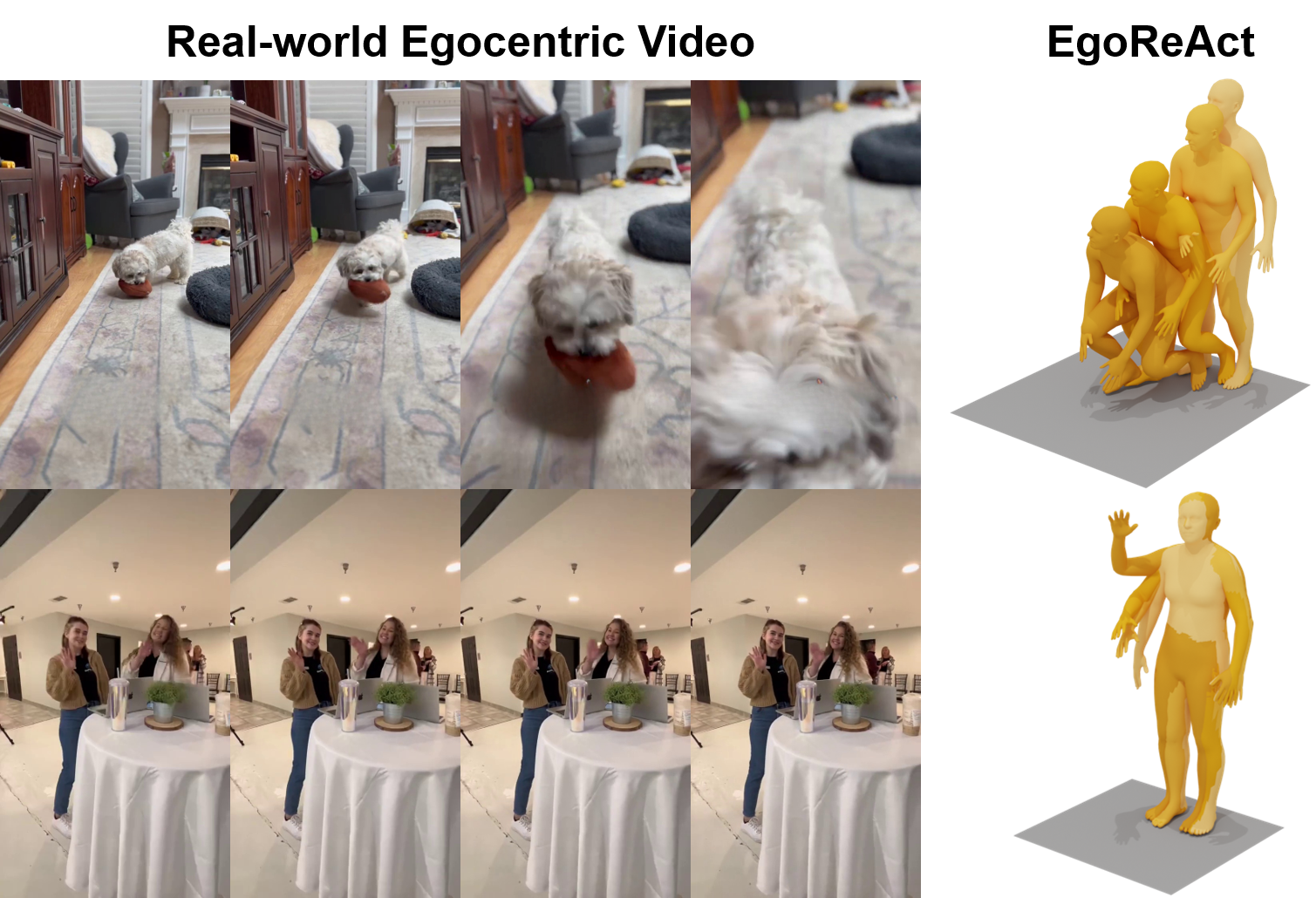}
    \vspace{-3mm}
\caption{\textbf{Real-world System Deployment.} The figure demonstrates the performance of our method on real-world first-person videos. On the left is the first-person video (selected from YouTube), while on the right are the human reactions generated by our method, EgoReAct, in response to these dynamic visual inputs.}   
    \label{fig:system_deployment}
    \vspace{-4mm}
\end{figure}

\begin{table}[t]
\centering
\small
\setlength{\tabcolsep}{3pt}
\renewcommand{\arraystretch}{1.3}
\resizebox{\linewidth}{!}{
\begin{tabular}{lcccc}
\toprule
\textbf{Method} & \textbf{FID}$\downarrow$ & \textbf{Diversity}$\rightarrow$ & \textbf{MModality}$\uparrow$ & \textbf{Head Traj Error (cm)}$\downarrow$\\
\midrule
Real & - & $8.098^{\pm0.054}$ & - & -\\
\midrule
\textit{w/o} Metric Depth & $0.619^{\pm0.013}$ & $7.772^{\pm0.064}$ & $2.520^{\pm0.048}$ & $69.2^{\pm0.424}$\\
\textit{w/o} Head Dynamics & $0.559^{\pm0.014}$ & $7.937^{\pm0.063}$ & $2.419^{\pm0.063}$ & $81.0^{\pm0.592}$\\
\textbf{Ours (full)} & $\textbf{0.456}^{\pm0.017}$ & $\textbf{8.042}^{\pm0.053}$ & $\textbf{2.581}^{\pm0.060}$ & $\textbf{65.7}^{\pm0.397}$\\
\bottomrule
\end{tabular}
}
\caption{\textbf{Ablation study on 3D dynamics.} We assess the impact of metric depth cues and head dynamics on model performance.}
\label{tab:ablation_3d}
\end{table}

\begin{table}[t]
\centering
\small
\setlength{\tabcolsep}{3pt}
\renewcommand{\arraystretch}{1.3}
\resizebox{\linewidth}{!}{
\begin{tabular}{lcccc}
\toprule
\textbf{Method} & \textbf{FID}$\downarrow$ & \textbf{Diversity}$\rightarrow$ & \textbf{MModality}$\uparrow$ & \textbf{Head Traj Error (cm)}$\downarrow$\\
\midrule
Real & - & $8.098^{\pm0.054}$ & - & -\\
\midrule
Ours (\textit{w/o} SA) & $0.931^{\pm0.011}$ & $7.767^{\pm0.042}$ & $2.430^{\pm0.059}$ & $71.6^{\pm0.459}$\\
\textbf{Ours (\textit{w/} SA)} & $\textbf{0.456}^{\pm0.017}$ & $\textbf{8.042}^{\pm0.053}$ & $\textbf{2.581}^{\pm0.060}$ & $\textbf{65.7}^{\pm0.397}$\\
\bottomrule
\end{tabular}
}
\caption{\textbf{Ablation study on data generation strategy.} Comparison of the model performance using the automated data generation pipeline with and without spatial alignment (SA) in the training data.}
\label{tab:ablation_dataset}
\end{table}

\subsection{Experimental Settings}
\noindent\textbf{Dataset Partitioning.}
After the data curation in~\cref{subsec: dataset}, we collect our dataset, \dbname, which has 3500 human reaction motion with spatial-aligned ego video.
We divide it into training and test sets with a ratio of 4:1, 
yielding 2,800 pairs for training and 700 pairs for testing.

\noindent\textbf{Evaluation Metrics.}
Following prior works~\cite{humanml3d, guo2020action2motion, xu2024regennet}, we adopt standard metrics widely used in human motion generation, 
including Frechet Inception Distance (FID), diversity, and multimodality.
To further evaluate spatial consistency, we report the global head trajectory error (Traj Error) which is the mean Euclidean distance between predicted and reference head trajectories in global space.
We also measure the First-Frame Latency (FFL) to compare the runtime efficiency of different generation methods.

\noindent\textbf{Implementation Details.}
For efficient inference, we use the DINO v2 small model for image encoding and the video-depth-anything small model for depth estimation. The autoregressive transformer, with a latent dimension of 1024, 6 attention heads, and 8 layers, 
is trained with a batch size of 64 for 100 epochs on a single NVIDIA H100 GPU, taking approximately 8 hours.

\subsection{Comparison to State-of-the-art Approaches}
To comprehensively evaluate our ego-perception–driven human reaction generation model, we conduct comparisons not only with HERO~\cite{yu2025hero}, the current state-of-the-art method specifically designed for video-conditioned reaction generation, but also with representative human motion generation models, MDM~\cite{tevet2022human} and EMDM~\cite{zhou2024emdm}.
Although MDM and EMDM were not originally developed for video-conditioned motion generation, we adapt them by replacing their text encoders with a DINOv2-based video encoder~\cite{oquab2023dinov2} to ensure consistent visual conditioning across methods.
All models are evaluated on the same dataset and under identical experimental settings to guarantee a fair comparison.

\noindent
\textbf{Quantitative Results}
As shown in Tab.~\ref{tab:comparison}, our method not only support fully-causally human reaction generation based on the input streaming video but also consistently outperforms all baselines across all metrics.
Benefited from causality and rich ego-perception representation, our model can generate more diverse and plausible human reaction motion based on different input video even they are in the similar scenarios.
Additionally, leveraging the spatial information, our model can synthesize more world-grounded motion based on the video input since the head trajectory is much close to the ground truth. 
Most importantly, our causal framework can support real-time streaming motion generation (\ie 45 FPS) on a single NVIDIA H100 GPU.

\noindent
\textbf{Qualitative Results.}
In Fig.~\ref{fig:Qualitative}, we provide side-by-side comparisons with HERO~\cite{yu2025hero}, MDM~\cite{Tevet2023HumanMotionDiffusion} and EMDM~\cite{zhou2024emdm} on four representative scenarios: \textit{AutoMobile\_Rush\_Towards}, \textit{Explosion}, \textit{Cat\_Come\_Towards}, and \textit{Tiger\_Come\_Towards}. Our method exhibits coherent global displacement and body orientation that remain tightly coupled to the egocentric stream, producing reactions that are temporally smooth and geometrically well grounded in the surrounding scene. In fast-approaching hazard cases (e.g., vehicle or tiger), our predictions show decisive retreat and turn-away behaviors synchronized with the camera’s advance, whereas baselines often display reduced translation magnitude, late responses, or noticeable drift relative to the camera path. In highly dynamic visual events (e.g., explosion), our outputs maintain stable foot contact and consistent motion amplitude, avoiding the over-conservative responses frequently observed in alternatives. For small, near-field triggers (e.g., cat approaching), our method yields fine-grained adjustments—subtle torso lean and step placement—while competing models either under-react or oscillate. These qualitative trends align with the quantitative improvements reported in Tab.~\ref{tab:comparison} and with human preference judgments, indicating superior spatial alignment, reaction plausibility, and overall motion quality.

\noindent
\textbf{User Study.} We invite 30 participants to compare reaction motions generated by different methods, including ours, HERO~\cite{yu2025hero},  MDM~\cite{Tevet2023HumanMotionDiffusion} and EMDM~\cite{zhou2024emdm}.  
Each participant selects the best motion for the following criteria: 
(1) Spatial Alignment: which demonstrates the strongest spatial alignment between the camera trajectory from egocentric video input and the human head trajectory? 
(2) Reaction Plausibility: which most plausibly matches the expected human reaction from the egocentric video? 
(3) Motion Quality: which exhibits the highest quality of body movement?



As shown in Fig.~\ref{fig:user_study}, participants preferred our method on all three criteria—\emph{Spatial Alignment}, \emph{Reaction Plausibility}, and \emph{Motion Quality}—selecting our results in $68.5\%$, $59.5\%$, and $58.6\%$ of trials, respectively, compared to at most $17.1\%$ for any baseline. The largest margin appears in \emph{Spatial Alignment} ($68.5\%$ vs.\ $16.2\%$ for HERO), indicating that our trajectories remain tightly coupled to the egocentric camera path, while gains in \emph{Reaction Plausibility} and \emph{Motion Quality} correspond to more context-appropriate responses and fewer motion artifacts (e.g., reduced foot sliding and jitter).

\subsection{Evaluation on Training Datasets}
We validate the effectiveness of our automated data generation strategy by training models on both the spatially aligned and original datasets, and evaluating them on the spatially aligned version.
As shown in ~\cref{tab:ablation_dataset}, models trained on the spatially aligned data achieve consistently better performance across all metrics compared to those trained on the original dataset.
This improvement can be attributed to the spatial alignment introduced by our pipeline, which ensures that the egocentric videos and human reaction motions are properly synchronized.
By correcting the misalignment in the original data, our approach enables the model to learn more consistent relationships between the video and the reaction motion, 
leading to better generalization and improved performance on downstream tasks.
Furthermore, the improved spatial alignment reduces the ambiguity in the motion generation process, 
allowing the model to produce more realistic and coherent human reactions.
\subsection{Ablation Study}

We conduct controlled ablations to disentangle the contribution of (i) metric depth cues, (ii) head/ego dynamics guidance. Unless otherwise noted, all training schedules, architectures, and evaluation splits remain identical to the full model; we report perceptual realism (FID), sample diversity/multimodality, and global head-trajectory error on the same test set.



\noindent\textbf{Effect of Depth.} 
Depth provides essential geometric cues for maintaining spatial consistency, and its absence leads to less accurate representations of the 3D scene, which in turn affects the overall quality. Without per-frame metric depth, the generator loses absolute scale and relative-distance signals that anchor the reaction to the 3D layout. 
Quantitatively, as shown in Tab.~\ref{tab:ablation_3d}, without per-frame metric depth, the generator loses absolute scale and relative-distance signals that anchor the reaction to the 3D layout. Removing the depth input causes a significant increase in FID, from 0.456 to 0.619, indicating a decrease in the visual realism of the generated motions. Besides, the Diversity score decreases from 8.042 to 7.772, suggesting that the lack of depth information limits the model's ability to generate diverse and varied motions. 
These results highlight the importance of depth in improving both the visual realism and diversity of the generated human reactions.

\noindent\textbf{Effect of Head Dynamics.} 
Suppressing the estimated head (ego) dynamics primarily harms global alignment and temporal causality. The synthesized body tends to “float” around the camera path, accumulating drift over time. Specifically, removing the head dynamics input results in a significant increase in multimodality, from 2.581 to 2.763, and a rise in Head Trajectory Error from 65.7 cm to 81.0 cm. 
This indicates that head dynamics are crucial for both maintaining the spatial alignment and ensuring the motion quality of the generated reactions. 
The multimodality increase reflects a reduced ability to match the generated motion with the diverse scenarios in the input video, while the larger Head Trajectory Error suggests that without head dynamics, 
the model struggles to maintain accurate alignment between the human head and the camera trajectory in the egocentric video.

Combining metric depth with head dynamics yields the best performance across all metrics. Depth supplies static scene geometry (scale, support, and occluder layout), while ego dynamics injects temporal constraints and camera–body coupling. Their synergy encourages reaction primitives that are jointly \emph{plausible} (low FID), \emph{expressive} (high diversity/multimodality), and \emph{geometrically consistent} (low trajectory error), validating our design choice.

\subsection{Real-world System Deployment}
To demonstrate the practical viability of our approach, we deploy EgoReAct on real-world egocentric video streams.
In contrast to synthetic datasets, real-world scenarios present various challenges such as unpredictable movement, dynamic backgrounds, and lighting changes, which require the model to be both flexible and efficient. 
As shown in \cref{fig:system_deployment}, our model effectively generates realistic human reactions in response to dynamic real-world events, maintaining both spatial alignment and temporal consistency throughout the video. 
This deployment demonstrates the robustness of our method in handling complex, real-world inputs, showing that despite being primarily trained on synthetic data, the model generalizes effectively and performs well in real-world settings, adapting to new, unseen scenarios.

\section{Conclusion}

\begin{figure}[t]
    \centering
    \includegraphics[width=0.45\textwidth]{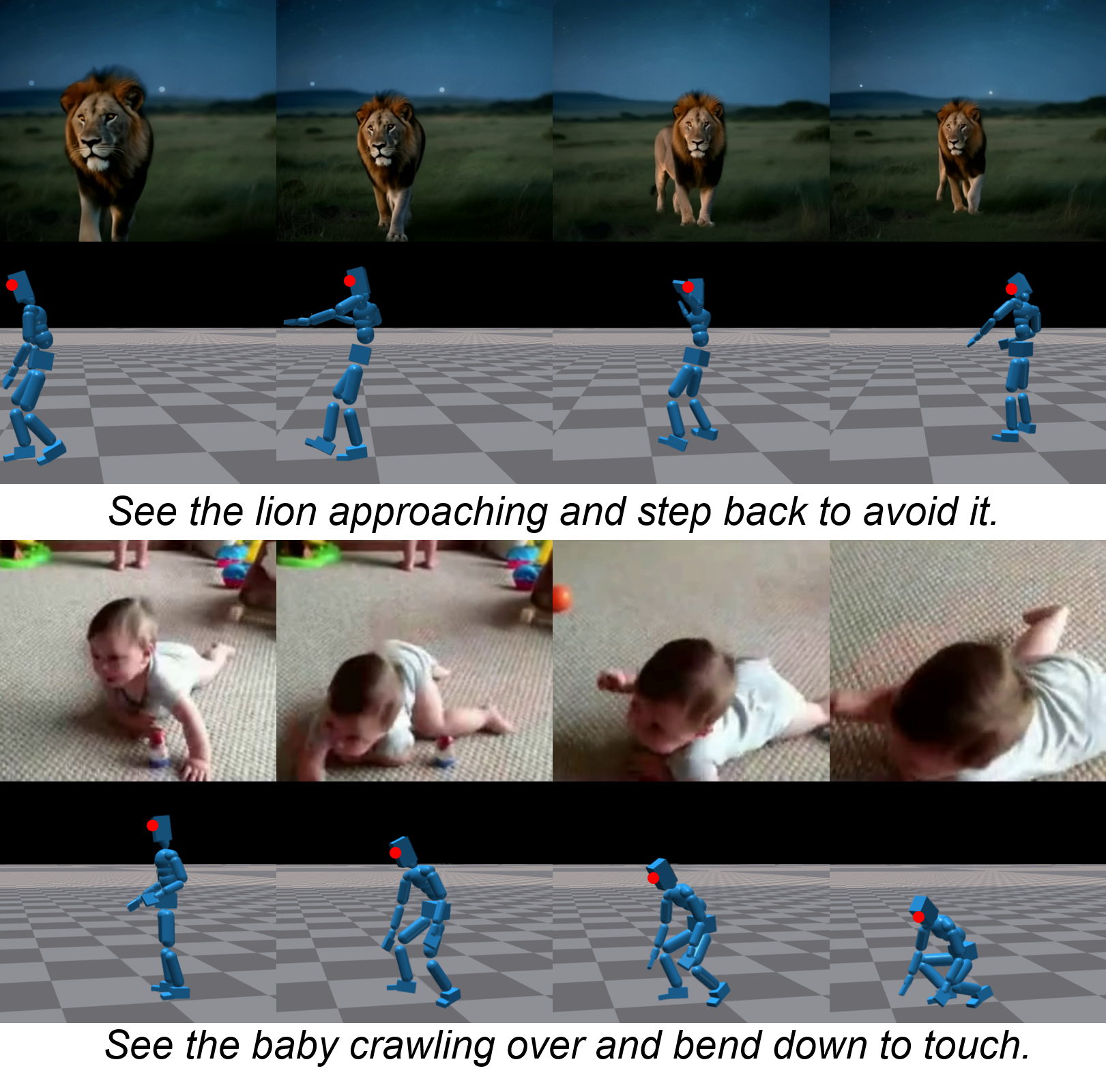}
        \vspace{-4mm}
\caption{\name takes egocentric video clips (first and third rows) as input and synthesizes temporally coherent 3D humanoid reactions (second and fourth rows) that respond to the observed events, serving as an effective high-level planner for sound- and language-guided humanoid behavior. The videos are captured from a first-person viewpoint, while the motions are visualized in a third-person view.
The humanoid’s viewpoint is annotated with a red point.}
    \label{fig:humanoid}
    \vspace{-4mm}
\end{figure}
We presented EgoReAct, a real-time autoregressive framework for generating spatially grounded human reaction motions from egocentric video streams.
By leveraging the newly constructed Human Reaction Dataset (HRD)—a spatially aligned egocentric video–reaction dataset built through an automated generation pipeline—
our approach provides a promising direction toward reducing data misalignment and scarcity in ego-centric reaction modeling.
EgoReAct models reactions in a strictly causal manner while incorporating 3D dynamic cues such as metric depth and head dynamics, achieving both geometric consistency and temporal coherence.
Extensive experiments demonstrate that our framework surpasses prior methods in spatial alignment, perceptual realism, and generation efficiency, establishing a foundation for future research in ego-driven human behavior modeling and embodied AI; See Fig.~\ref{fig:humanoid}.

\noindent\textbf{Limitations and Future Work.} 
While our framework effectively generates spatially grounded human reaction motions in real time, several limitations remain. 
First, although it leverages video diffusion models~\cite{alibaba_wan_2025, kuaishou_kling_2024} for scalable data generation, 
video quality could be further improved, particularly in fine-grained motion details and artifact reduction. 
Second, the physical plausibility of the generated motions can be enhanced. 
Our system currently focuses on kinematics-based reaction synthesis, ensuring that the reactions adhere more closely to physical realism~\cite{dou2023c, luo2023perpetual, huang2025modskill, zhang2024physpt, pan2025tokenhsi, wang2024sims} is an important future direction.
\clearpage
{
    \small
    \bibliographystyle{ieeenat_fullname}
    \bibliography{main}
}


\end{document}